\documentclass[runningheads]{llncs}

\usepackage{eccv}

\usepackage{eccvabbrv}

\usepackage{graphicx}
\usepackage{booktabs}

\usepackage[accsupp]{axessibility}  

\usepackage{hyperref}

\usepackage{orcidlink}

\usepackage{pifont}  
\usepackage{xcolor}  
\usepackage[dvipsnames]{xcolor}
\usepackage{textcomp, gensymb}

\newcommand{\cmark}{\textcolor{green}{\ding{51}}}  
\newcommand{\xmark}{\textcolor{red}{\ding{55}}}  

\begin{document}

\title{Upper-Body Pose-based Gaze Estimation for Privacy-Preserving 3D Gaze Target Detection} 

\titlerunning{Privacy-Preserving 3D Gaze Target Detection}

\author{Andrea Toaiari\inst{1}\orcidlink{0000-0002-3759-8865} \and
Vittorio Murino\inst{2}\orcidlink{0000-0002-8645-2328} \and
Marco Cristani\inst{1,3}\orcidlink{0000-0002-0523-6042} \and
Cigdem Beyan\inst{2}\orcidlink{0000-0002-9583-0087}}

\authorrunning{A. Toaiari et al.}

\institute{
Dept. of Engineering for Innovation Medicine, University of Verona, Verona, Italy
\and 
Dept. of Computer Science, University of Verona, Verona, Italy 
\email{name.surname@univr.it}\\
\and
QUALYCO S.r.l., Spin-off of the University of Verona, Verona, Italy\\
\email{marco.cristani@qualyco.com}
}

\maketitle

\begin{abstract}
  Gaze Target Detection (GTD), \ie, determining where a person is looking within a scene from an external viewpoint, is a challenging task, particularly in 3D space. Existing approaches heavily rely on analyzing the person's appearance, primarily focusing on their face to predict the gaze target.
  This paper presents a novel approach to tackle this problem by utilizing the person's upper-body pose and available depth maps to extract a 3D gaze direction and employing a multi-stage or an end-to-end pipeline to predict the gazed target.
  When predicted accurately, the human body pose can provide valuable information about the head pose, which is a good approximation of the gaze direction, as well as the position of the arms and hands, which are linked to the activity the person is performing and the objects they are likely focusing on. 
  Consequently, in addition to performing gaze estimation in 3D, we are also able to perform GTD simultaneously.
  We demonstrate state-of-the-art results on the most comprehensive publicly accessible 3D gaze target detection dataset without requiring images of the person's face, thus promoting privacy preservation in various application contexts.
  The code is available at \url{https://github.com/intelligolabs/privacy-gtd-3D}.
\end{abstract}

\section{Introduction}
\label{sec:intro}
Human communication relies on a complex interplay of verbal and non-verbal cues. Among nonverbal cues, the gaze holds particular significance as it reveals where a person directs their visual attention, offering insights into their interests, intentions, or imminent actions~\cite{emery2000eyes}. This ability to interpret gaze is crucial for understanding the reciprocal focus and facilitating interactions like turn-taking in social contexts~\cite{beyan2023co}.

Gaze analysis spans a broad spectrum of disciplines, including, among others, human-computer interaction, neuroscience, social robotics, and organizational psychology~\cite{noureddin2005non,dalton2005gaze,admoni2017social,capozzi2019tracking}. 
An analytical approach to estimating the gaze direction or the location which is gazed at not only enhances our understanding of human behavior but also informs the development of technologies that rely on human-machine interaction, social understanding in robotics, and even psychological research into interpersonal dynamics and attentional processes~\cite{rayner1998eye,edwards2015social,beyan2019sequential}. 
Thus, the study of gaze represents a multidisciplinary endeavor that continues to evolve with technological and cognitive science advancements.

In computer vision, the study of gaze behavior primarily involves two main tasks: \emph{gaze estimation} and \emph{gaze target detection}.
Gaze estimation typically implicates predicting where a person is looking in a three-dimensional space. This means determining the direction of a person's gaze in terms of yaw, pitch, and sometimes roll angles, which correspond to horizontal, vertical, and depth dimensions, respectively. 
Gaze estimation can be performed from a single image or a sequence of images, often focusing on the person's eyes or the entire head region~\cite{Cheng_2018,guo2020domain,kellnhofer2019gaze360}.
Gaze target detection~\cite{recasens2015they} (also referred to as gaze-following), on the other hand, focuses on identifying the specific point or region in the scene that the person is looking at. This is typically done in a two-dimensional space (2D), where the goal is to localize the gaze target's exact coordinates $(x, y)$ within the image or video frame.

\begin{figure}[!tb]
  \centering
  \includegraphics[width=0.9\linewidth]{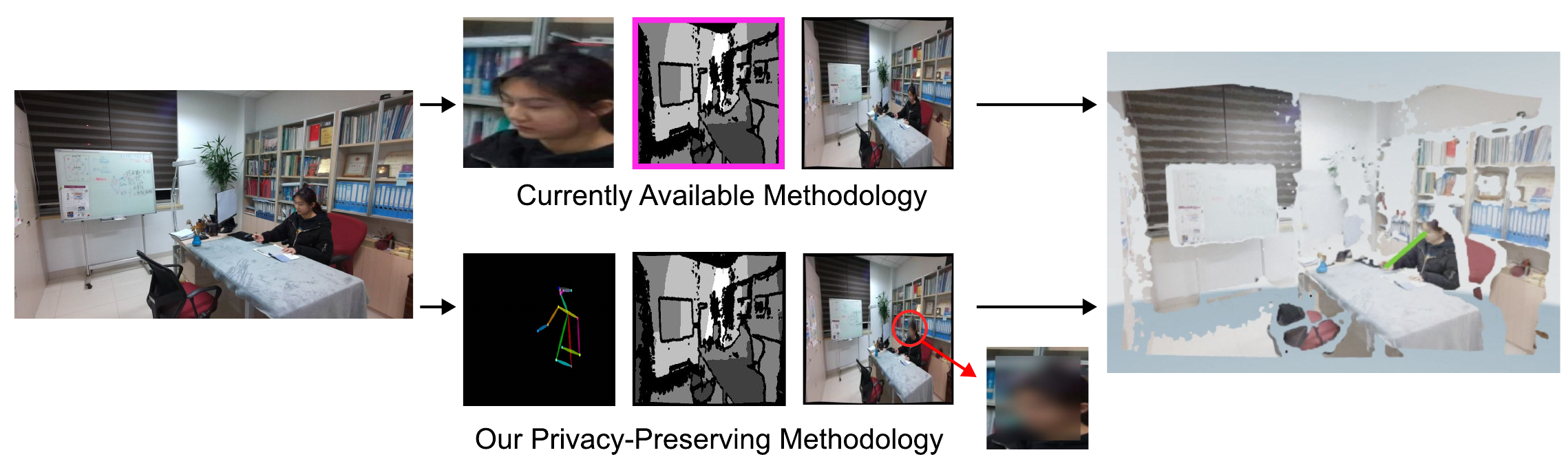}
  \caption{We introduce a novel way to tackle the 3D gaze estimation and 3D gaze target detection problems that, differently from the previous approaches, does not use the head crops of the observed people but rather employs upper-body skeletons, depth maps, and scene images where the head is blurred out. 
  The new pipeline not only proved successful in tackling both tasks but offers a concrete way to preserve the identity of the involved people.
  The currently available state-of-the-art method already used depth maps (in {\color{magenta}magenta}) in its pipeline, so we are not introducing additional information. 
  }
  \label{fig:teaser}
\vspace{-2em}
\end{figure}

Traditional methods for gaze estimation historically relied on geometric eye models and regression functions to establish a mapping from eye or face images to gaze vectors~\cite{lu2016estimating,nakazawa2012point,valenti2011combining}. These methods demonstrated effectiveness in controlled environments with consistent subject characteristics, head positions, and lighting conditions; however, their performance diminishes in more diverse and less controlled settings~\cite{zhang2019evaluation}.
In recent years, some methods incorporated broader facial features such as head pose and overall facial appearance to enhance accuracy and indicated that leveraging full-face information tends to yield more precise gaze estimates compared to methods focusing solely on eye images~\cite{zhang2017s,krafka2016eye}. To the best of our knowledge, no research has explored the use of full-body or upper-body pose rather than head orientation or head/eye images for \textbf{3D gaze estimation} in unconstrained scenes.

When addressing \textbf{gaze target detection}, the predominant approach has focused on performing this task within a 2D context, utilizing a dual-pathway deep architecture. This architecture typically incorporates inputs such as head crops and scene images~\cite{chong2018connecting,lian2018believe,recasens2015they,Recasens2017}.
Some studies have expanded upon this framework by introducing temporal data integration \cite{chong2020detecting} or by incorporating a third pathway to handle depth maps of the scene \cite{fang2021dual, jin2022depth, miao2023patch, tonini2022multimodal}.
While numerous attempts have been made to perform gaze target detection in 2D, significantly fewer studies have focused on 3D gaze target detection. 
However, it has been demonstrated that gaze following in a 3D scene with only a single camera is feasible \cite{wei2018and,brau2018multiple,hu2023gfie}. 

\noindent\textbf{Our proposal.}
In this paper, we propose a methodology that not only solves the \textbf{3D gaze estimation task} in a novel way, employing estimated upper-body skeletons and ground truth depth maps, but also provides a solution for both \textbf{2D and 3D gaze target detection} while simultaneously preserving the identity of the observed subjects.
Inspired by the preliminary work carried out in~\cite{gupta2022modular}, we build upon the framework and the dataset proposed by~\cite{hu2023gfie}, replacing their gaze estimation module based on head crops with a skeleton and depth-based module, limiting the treatment of scene and face images only to the pose estimation pre-processing phase, after which all these sensitive information can be discarded (see Fig.~\ref{fig:teaser}).
A single attention layer, applied to the combination of \textbf{upper-body pose} features normalized on one anchor joint and convolutional features extracted from the scene depth maps, provided us with accurate 3D gaze vectors.
These vectors are then fed to a 3D gaze target estimation pipeline, which uses scene images where the people's faces have been blurred out.
The proposed method can be trained either in a \textbf{multi-stage} fashion, pre-training the gaze estimation module, or in an \textbf{end-to-end} manner. 
This approach not only achieves state-of-the-art results on the GFIE dataset \cite{hu2023gfie}, the most comprehensive dataset available for 3D scenarios but also represents a novel approach to privacy preservation in the context of 3D gaze target detection. 

\noindent\textbf{Our contributions.}
The main contributions of our work can be summarized as follows:
\begin{itemize} 
    \item We present a novel module for \textbf{3D gaze estimation} that utilizes upper-body skeleton data and the depth map of the scene. This approach achieves outstanding results, showcasing the richness of pose data in understanding a person's gaze direction in 3D.
    \item We incorporate the above-mentioned new module into a pipeline for \textbf{3D gaze target estimation}, achieving state-of-the-art results on the GFIE dataset \cite{hu2023gfie}, the most comprehensive dataset currently available for the task at hand.
    \item For the first time, we focus on the \textbf{privacy-preserving} capabilities of a vision-based gaze estimation and gaze target detection approach in the 3D space, discarding information about the person's facial appearance and concentrating solely on their upper-body poses. 
    By doing so, our approach eliminates the need to use a head backbone pre-trained on a large dataset of head orientations.
\end{itemize}
\section{Related Work}
\label{sec:related}
We provide an overview of the standard gaze target detection task, usually defined in the 2D space, in Section~\ref{subsec:gtd}, and the recent developments when it comes to tackling it in the 3D space in Section~\ref{subsec:gtd-3d}.
Finally, in Section~\ref{subsec:ge}, we offer a brief panoramic of the related gaze estimation task.

\subsection{Gaze Target Detection in 2D}
\label{subsec:gtd}
Gaze target detection is a complex vision task that consists of finding what an observed human subject is looking at, usually by analyzing only an image or a video of the scene taken from a third-person point of view. 
\cite{recasens2015they} formalized it, proposing a 2D dataset with multiple target annotations and a double pathway model that extracts features from a subject's head appearance and the scene's salient areas.

This structure has been adopted by multiple following works~\cite{chong2018connecting,lian2018believe,recasens2015they,Recasens2017,chong2020detecting}. 
For instance,~\cite{chong2020detecting} proposed a spatiotemporal modelling of the gaze target detection problem and a way to classify whether the gaze target was inside or outside the scene image. 
Among these, some introduced a third pathway to incorporate the depth map of the scene image, which is derived using a monocular depth estimator~\cite{fang2021dual,jin2022depth,miao2023patch,tonini2022multimodal}.
While the approaches outlined in~\cite{recasens2015they,chong2020detecting,fang2021dual,jin2022depth,miao2023patch,tonini2022multimodal} have proven effective, they are unable to simultaneously detect gaze targets for all the individuals within a scene. 
This limitation is addressed by Transformer-based gaze target detectors, as shown in~\cite{tu2022end,Tonini_2023_ICCV}. On the flip side, a notable drawback of~\cite{tu2022end,Tonini_2023_ICCV} is their requirement for relatively larger training datasets, causing them to underperform in a low-data regime. 

The work of~\cite{gupta2022modular} presented the possibility of using skeleton data and depth maps, but their method underperformed compared to earlier work. 
Note that their implementation uses skeleton data represented as images, whereas our pipeline utilizes normalized key points (\ie, not images).

\subsection{Gaze Target Detection in 3D}
\label{subsec:gtd-3d}
Although various efforts have been made to extend 2D gaze target detection to the 3D space~\cite{bao2022escnet,horanyi2023they} by estimating depth maps and three-dimensional gaze cones, these methods have been tested only on datasets~\cite{recasens2015they,chong2020detecting} that provide 2D data and annotations. 
Similarly, the estimation of visual selective attention~\cite{toaiari2023scene} involves mapping attention directly onto the 3D point cloud of a scene by analyzing video sequences of individuals moving within a constrained environment. This approach uses 2D inputs but performs its evaluation entirely in 3D space, focusing on how multiple subjects' visual attention maps onto the scene over time.
Wei et al.~\cite{wei2018and} is one of the first methods to deal with this task in the 3D scenario, addressing not only the attention prediction challenge but also investigating why the subject is looking there and what action is being performed.

In \cite{hu2022we}, an RGB-D dataset with 3D annotations has been proposed, offering manually annotated 3D gaze targets and adding a novel 3D pathway to the now standard saliency-head prediction pipeline.
Recently, Hu et al.~\cite{hu2023gfie} presented an architecture with four modules for a) estimating gaze directions, b) perceiving the stereo field of view, c) generating a 2D gaze heatmap, and d) retrieving the 3D gaze target. This method utilizes a pre-trained ResNet50 as the backbone for the gaze direction estimation module, which takes a cropped head image as input and outputs a 3D gaze unit vector. From the 2D prediction, the 3D target is then retrieved in the evaluation phase. A dataset called GFIE is also presented in the paper, representing the most comprehensive data collection to date for gaze target detection in 3D.

\subsection{Gaze Estimation}
\label{subsec:ge}
The field of computer vision has a longer history in gaze estimation compared to gaze target detection. 
There are two main approaches to gaze estimation: geometric eye models and appearance models. The latter has become more prevalent with the rise of deep learning. 
The most recent appearance-based models have shown increased interest in CNN-based solutions. For instance, Zhang et al.~\cite{zhang2017mpiigaze} use both eye and facial features, while Krafka et al.~\cite{krafka2016eye} developed a multi-channel network using eye images, full-face images, and face grid information. 
Chen et al.~\cite{chen2022towards} introduced a model integrating statistical information within a CNN for eye images, utilizing dilated convolutions to retain high-level image features without reducing spatial resolution. Their GEDDNet model employs gaze decomposition with dilated convolutions for enhanced performance.

Fischer et al.~\cite{fischer2018rt} enhanced gaze prediction accuracy by incorporating the head pose vector with VGG CNN-extracted features from eye crops and using an ensemble scheme.
Cheng et al.~\cite{cheng2020coarse} proposed CA-Net, which first predicts primary gaze angles from a face image and refines them with residuals from eye crops, using a bi-gram model to integrate primary gaze and eye residuals. 
Kellnhofer et al.~\cite{kellnhofer2019gaze360} employed an LSTM-based temporal model with 7-frame sequences and adapted pinball loss for joint gaze direction and error bounds regression to improve accuracy.

More recently, transformers with self-attention modules have been applied to gaze estimation. One approach~\cite{o2022self} used a transformer to extract crucial gaze features from images with high variance, filtering irrelevant information using convolution projection and preserving detailed features with a deconvolution layer. AGE-Net~\cite{biswas2021appearance} proposed using two parallel networks for each eye image: one generating a feature vector with a CNN and the other generating a weight feature vector with an attention-based network.
Notably, no existing work addresses processing body or upper-body pose and depth maps for gaze estimation in 3D, as we target in this study. 
Furthermore, the literature focusing on both gaze estimation and 2D/3D gaze target detection simultaneously (\eg,~\cite{hu2023gfie}), which is our aim, is very limited. For those interested in a deeper exploration, survey papers on gaze estimation are available for reference~\cite{ghosh2023automatic,pathirana2022eye}.
\section{Methodology}
\label{sec:method}

In this section, we introduce our upper-body and depth-based gaze estimation module and explain how it enables us to execute the entire 3D gaze target detection pipeline while preserving the privacy of the involved subjects.
The proposed architecture is presented in Fig.~\ref{fig:approach} and described in Section~\ref{subsec:pipeline}.

Section~\ref{subsec:gaze-estimation} describes our novel gaze estimation method and how it fits inside the gaze target detection pipeline.
Privacy preservation and how it is enforced is discussed in Section~\ref{subsec:privacy}, while in Section~\ref{subsec:training} and Section~\ref{subsec:implementation} we provide training and implementation details to enable appropriate reproducibility of the proposed approach.

\subsection{Gaze Target Detection Pipeline}
\label{subsec:pipeline}
Given an image or video dataset which offers annotations about the ground-truth gaze vector and the 2D/3D gaze targets, the task of 3D gaze target detection consists of finding the 3D point $t^{3D} = (t^{3D}_x, t^{3D}_y, t^{3D}_z)$ looked at by the subject, or subjects if extended to social gaze prediction, in the scene and their gaze direction represented by the normalized vector $g =(g_x,g_y,g_z)$.
The input data for each gaze prediction includes the image scene $s$, the aligned depth map $d$ with associated camera parameters $k$, a binary mask for the face location $m$, the 2D ($e^{2D}$) and 3D ($e^{3D}$) annotations of the point between the eyes and the bounding box ($box_h$) around the head. 
Note that the head bounding box is not employed as a modality but is used to blur the individual's face in the scene image; 
For further details, see Section~\ref{subsec:privacy}.

\begin{figure}[!tb]
  \centering
  \includegraphics[width=0.95\linewidth]{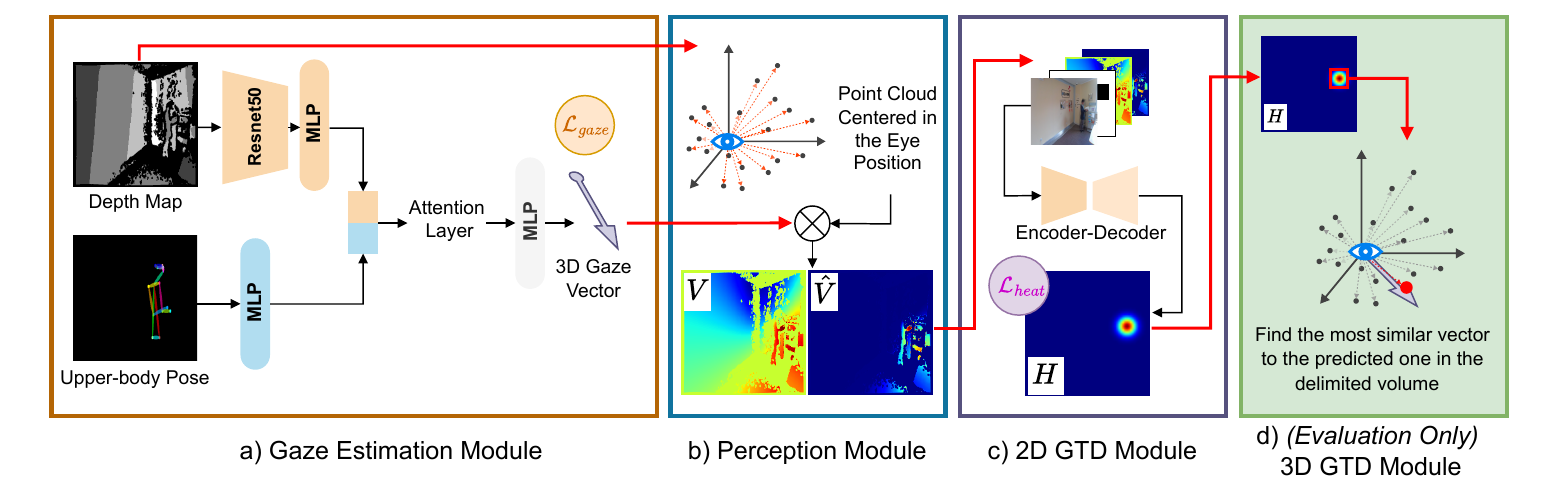}
  \caption{Overview of the proposed approach. \textbf{a)} The 3D Gaze Estimation Module predicts a gaze vector by exploiting the upper-body pose coordinates, processed by a simple MLP, and the convolutional features extracted from the depth map by a ResNet50~\cite{he2016deep}. A single multi-head attention layer is applied to the concatenated features before the final MLP.
  \textbf{b)} The perception module converts the depth map to the unprojected point cloud, from which the 3D eye coordinates are subtracted, and computes two heatmaps highlighting the most interesting part of the scene. 
  \textbf{c)} The 2D gaze target heatmap is predicted by an encoder-decoder architecture operating on the concatenation of the scene image (with the blurred-out face), the head location mask and the two heatmaps. 
  \textbf{d)} During the evaluation, the most similar vector to the predicted vector $\hat{g}$ in the unprojected and translated point cloud is selected as the final gaze vector and the corresponding point in the original point cloud as the predicted 3D gaze target.}
  \label{fig:approach}
  \vspace{-1em}
\end{figure}

We adopted the four-module pipeline presented in~\cite{hu2023gfie} to extract the 2D and the 3D gaze targets from the 2D image and the 3D point cloud, respectively. 
The gaze estimation module, Fig.~\ref{fig:approach}a, computes the normalized three-dimensional gaze vector $\hat{g} =(\hat{g_x}, \hat{g_y}, \hat{g_z})$, starting from the annotated 3D eye position ${e^{3D} = (e^{3D}_x, e^{3D}_y, e^{3D}_z)}$ and describing the direction of where the person in the scene is looking.
The second module (Fig.~\ref{fig:approach}b) creates a heatmap $V$ defined as the dot product between the unprojected depth map $d' - e^{3D}$, corresponding to the point cloud of the scene centered in the 3D eye position $e^{3D}$, and the estimated gaze vector $\hat{g}$, with the maximum values representing the points in the point cloud with the highest probability of being looked at. 
Intuitively, each unprojected (and translated) point represents a vector from $e^{3D}$ to that point, and the heatmap role is to highlight the closest vectors to the estimated gaze direction.
A second heatmap $\hat{V}$ is computed as $\hat{V} = ReLU(V)^\alpha$, with $\alpha = 3$, to further highlight the most interesting points in the scene.
The scene image $s$, the heatmaps $V$ and $\hat{V}$, and the head location mask $m$ are then concatenated and fed to the third module (Fig.~\ref{fig:approach}c), an encoder-decoder network, that generates the 2D gaze heatmap $H$ used to extract the 2D gaze target. 

The last module, depicted in Fig.~\ref{fig:approach}d, is used only in the evaluation phase and is responsible for extracting the 3D gaze target from the original point cloud. 
After extracting the 2D gaze target from the predicted heatmap $H$ as \mbox{$t^{2D}=\texttt{argmax}(H)$}, a 3D volume centered in that point is built in the unprojected (and translated to the $e^{3D}$ position) vector space to enclose a portion of the available points. 
The vector with the highest similarity to the predicted gaze vector inside the volume is selected as the new predicted gaze vector $\hat{g}$, and the corresponding point
is selected as the 3D gaze target from the original point cloud. Below, we discuss the critical ingredients and the processing of our approach and, as a result, the novelty of the pipeline in detail.

\subsection{The Skeleton-based Gaze Estimation Module}
\label{subsec:gaze-estimation}
To effectively obtain a pipeline able to preserve the identity of the subjects and avoid using image data related to the appearance of the face, the first step we adopt consists of replacing the original gaze estimation method based on face crops~\cite{hu2023gfie}.
We, therefore, introduce a novel module for gaze estimation, \ie, the prediction of the gaze direction in the 3D space, that exploits the upper-body skeleton and the depth map as multi-modal information to extract a coherent gaze vector.
Considering the pose estimation as a pre-processing step, after which the whole gaze target detection pipeline doesn't require the face appearance, the proposed gaze estimation module has proven effective in correctly estimating the gaze direction despite having no access to either the face crop or the scene image.
The pose key points used in this study are from the hip upwards, for a total of 13 joints, including the arms and wrists, which are extracted using a pre-trained top-down pose estimator~\cite{yuan2021hrformer}. 

The key points go through a normalization procedure that uses the neck position and the distance between the neck and hip joints to obtain a scale-invariant representation of the human pose. 
A simple MLP with three layers processes them to expand the features' dimensionality. Concurrently, a convolutional encoder extracts the depth features from the resized depth map. 
The upper-body and depth features are concatenated, and a single attention layer is applied to select the most helpful features for predicting the gaze direction.
Finally, two linear layers are applied to these features to extract the 3D gaze vector $\hat{g}$, which is normalized before being handed over to the next module.

\subsection{Privacy Preservation}
\label{subsec:privacy}
The only step of the proposed pipeline where the face images are used is the pose estimation pre-processing phase, ensuring that the gaze target detection architecture can offer the claimed privacy-preserving feature.
In addition to the use of upper-body poses and depth maps to replace the need to use head crops, the entire scene image used as one of the inputs by the auto-encoder module for the prediction of the heatmap $H$ has been edited by applying Gaussian blur in the region defined by the head bounding box provided in the dataset, $box_h$, hence making sure that no module could use that information to improve the gaze target detection. 
Examples of edited scene images and skeletons extracted by the pose estimation method are shown in Fig.~\ref{fig:gfie-qualitatives}. 

\subsection{Training}
\label{subsec:training}
Excluding the pose estimation phase, where a pre-trained 2D pose estimator has been adopted, our gaze target detection pipeline can be trained in two ways: a) multi-stage training, and b) end-to-end manner.
For the multi-stage training, firstly, the gaze estimation module is trained standalone using only a cosine loss on the estimated gaze vector ${\mathcal{L}_{gaze} = 1 - CosineSimilarity(g, \hat{g})}$ on the same train-valid-test split as the complete architecture.
The module weights are then loaded into the same module, but this time as part of the full pipeline. The latter is then trained in an end-to-end fashion and supervised on the gaze vector estimation with a cosine loss $\mathcal{L}_{gaze}$ and on the 2D heatmap $H$ creation with an L2 loss $\mathcal{L}_{heat}$. The total loss is then computed as $\mathcal{L} = w_{heat}\mathcal{L}_{heat} + w_{gaze}\mathcal{L}_{gaze}$, where $w_{heat}$ and $w_{gaze}$ are two loss-specific weights defined as hyperparameters.
Specifically, $w_{heat}$ has a value of 10000, and $w_{gaze}$ has a value of 10, similar to \cite{hu2023gfie}'s implementation.
The second training approach, the so-called end-to-end, involves training the entire gaze target detection pipeline, including the gaze estimation module, in a unified manner. This method does not rely on any pre-trained gaze estimation module from either a different dataset or the training split of the dataset used.

\subsection{Implementation Details}
\label{subsec:implementation}
The proposed approach modules are implemented in PyTorch and trained on a machine equipped with an Intel-i9 CPU and a single NVIDIA RTX 4090. 
\\

\noindent\textbf{Gaze estimation module.} The module that manages the estimation of the gaze direction processes depth maps resized to $224\times224$ pixels using a ResNet50\cite{he2016deep} and three linear layers with ReLU and Dropout. A three-layer Multi-Layer Perceptron (MLP) with ReLU activation function handles the $13\times2$ upper-body pose data estimated from a pre-trained HRFormer~\cite{yuan2021hrformer} and normalized on the neck position and the distance between the neck and hip joints.
The attention layer is implemented as a Transformer Encoder Layer with four heads and a feature size of $256$.
In the multi-stage training regime, the gaze estimation module is trained standalone with a batch size of 128 for 30 epochs, the Adam~\cite{kingma2014adam} optimizer and a learning rate and weight decay of $1e-4$.
Given that prior work employs auxiliary large gaze estimation datasets for pre-training their gaze estimation module, we take a different approach for our multi-stage training. 
Instead of using an additional dataset, we utilize the weights obtained from training our gaze estimation module solely on the training set of the GFIE dataset~\cite{hu2023gfie} with the gaze loss $\mathcal{L}_{gaze}$ (corresponding results are given in Section~\ref{subsec:gaze-results}) to initialize the module within the complete gaze target detection pipeline. 

\noindent\textbf{Gaze target detection pipeline.} The full pipeline is trained for 50 epochs, with the Adam~\cite{kingma2014adam} optimizer, a batch size of 32, and a learning rate and weight decay of $1e-4$. 
The images and depth maps are resized to $224\times224$ pixels. 
Random crop, flip, saturation, brightness and contrast adjustments have been applied as augmentations. 
The predicted perception heatmaps $V$ and $\hat{V}$ have a size of $224\times224$ pixels, while the final 2D target detection heatmap $H$ has a size of $64\times64$ pixels. The ground truth heatmap is generated by a Gaussian centered in the gaze target~\cite{recasens2015they}. The encoder is implemented as a ResNet50~\cite{he2016deep}, pre-trained on Imagenet~\cite{deng2009imagenet}, while the decoder consists of two convolutional layers and three deconvolutional layers.
\section{Experiments}
\label{sec:exp}
In this section, we describe the dataset used (Section~\ref{subsec:dataset}) for our study, detailing the evaluation setup along with the 2D and 3D metrics used for gaze estimation and gaze target detection (Section~\ref{subsec:metrics}). We then showcase the results of our gaze estimation module (Section~\ref{subsec:gaze-results}), comparing them to the most related work~\cite{hu2023gfie}.
Next, we report the results of 2D/3D gaze target detection (Section~\ref{subsec:results}), corresponding to the entire pipeline and comparing our findings against state-of-the-art methods. Finally, we present the qualitative results (Section~\ref{subsec:qual}) and ablation study (Section~\ref{subsec:ablation}).

\subsection{Dataset}
\label{subsec:dataset}

Datasets annotated for 3D gaze estimation and 2D/3D gaze target detection are particularly scarce. Consequently, we use the largest available dataset of this kind, GFIE~\cite{hu2023gfie}, to evaluate our proposed method. This dataset allows us to compare our results with state-of-the-art methods and conduct an ablation study to examine the impact of using only a portion of the skeleton and the effect of blurring the face.

The GFIE dataset~\cite{hu2023gfie} was captured by using a laser rangefinder and an Azure Kinect RGB-D camera mounted on a stable platform. The laser rangefinder projects a laser spot to guide the subject's gaze, ensuring continuous fixation during recording. This way, the distance measurements from the laser rangefinder can be logged. Annotations encompass a) head bounding boxes, b) 2D/3D gaze targets, and c) 2D/3D eye locations. The annotations were performed semi-automatically. For example, a face detection method~\cite{zhang2017s3fd} was used, followed by manual verification. For 2D gaze targets, after detecting laser spots in RGB images, a team of annotators validates the accuracy of these detections. Using the distance measurements obtained from the laser rangefinder, the positions of the gaze targets in 2D are mapped into corresponding 3D coordinates in the real-world space. Additionally, using a facial landmark detector, the locations of left and right eye landmarks were identified within cropped head images. 
The 2D eye position corresponds to the midpoint between these landmarks, while the 3D eye position was computed by unprojecting from the depth map's facial landmarks.
This dataset includes diverse gaze behaviors from 61 subjects (27 male and 34 female), covering various activities. It consists of a total of 71,799 frames. The dataset is divided into a training set containing 59,217 frames, a test set with 6,281 frames, and a validation set containing 6,281 frames. 
Importantly, the subjects and scenes present in the training set were excluded from both the test and validation sets. 
To validate our approach, we use the same evaluation protocol defined in~\cite{hu2023gfie}, and during training, we use the annotations for the 2D and 3D gaze targets and the annotated 3D gaze vector as the ground truth.

\subsection{Evaluation Metrics}
\label{subsec:metrics}
While our primary focus remains on the 3D evaluation metrics and the 3D scene in general, we also report the results for the 2D metrics, usually adopted for the standard gaze target detection task.
The standard 3D metrics involve distance and the angle error defined as follows. \textbf{3D Distance} measures the Euclidean distance between the predicted and actual gaze points in a 3D space. The unit is meters. \textbf{Angle Error} quantifies the angle difference between the predicted and actual gaze directions. It is computed as the \texttt{arccos} of the cosine similarity between the normalized ground truth and predicted gaze vectors. The unit is degrees.
On the other hand, 2D metrics include the area under the curve and the 2D distance, described as follows. \textbf{Area Under Curve (AUC)} utilizes the predicted heatmap to plot the ROC curve, as proposed by~\cite{judd2009learning}. \textbf{2D Distance} computes the Euclidean distance between the predicted and actual gaze points, using the \texttt{argmax} of the predicted 2D heatmap, assuming an image size of 1$\times$1.
While angle error is a metric for 3D gaze estimation, all other metrics pertain to 2D/3D gaze target detection tasks.

\subsection{Results of the Gaze Estimation Module}
\label{subsec:gaze-results}
Without injecting our proposed module for gaze estimation into our complete gaze target detection pipeline, we trained and tested it alone. We performed the same for the face crop-based approach in~\cite{hu2023gfie}.
Differently from our module, Hu et al.~\cite{hu2023gfie} pre-trained their network that gets the head crops as input on another gaze estimation dataset called Gaze360~\cite{kellnhofer2019gaze360}. 
However, we do not use any auxiliary datasets to pre-train our model.
\cite{hu2023gfie}, while performing very well on the training set with a 3D angle error close to 3\degree, when it comes to the testing set, it achieves an angle error of 19.7\degree, showing an overfitting characteristic. On the other hand, our training and testing performance is consistent, and we perform a 3D angle error of 16.5\degree, which is not only remarkably better but also overfitting-free.

\subsection{Results on the GFIE Dataset}
\label{subsec:results}

The results obtained by running our full pipeline on the GFIE dataset~\cite{hu2023gfie} are reported in Table~\ref{tab:results}. We compare our method with several approaches for detecting gaze targets in both 2D and 3D, as summarized below.
\begin{itemize}
    \item \emph{Random} approach refers to gaze targets chosen randomly within the image for 2D and within the point cloud for 3D space.
    \item \emph{Center} approach positions the gaze targets at the center of the image for 2D space and at the center of the point cloud for 3D space.
    \item For the existing methods that focus on gaze target detection in 2D, such as \emph{GazeFollow}~\cite{recasens2015they}, \emph{Lian et al.}~\cite{lian2018believe}, and \emph{Chong et al.}~\cite{chong2020detecting}, they have been extended to estimate 3D gaze targets by unprojecting the 2D gaze targets into 3D space in~\cite{hu2023gfie}. 
    Additionally, \emph{RT-Gene}~\cite{fischer2018rt} and \emph{Gaze360}~\cite{kellnhofer2019gaze360}, which estimate 3D gaze directions in unconstrained environments, have been adapted to the same evaluation protocol in~\cite{hu2023gfie}. The most similar vector to the predicted gaze vector within the unregistered depth map has been selected as the actual gaze vector, and the point obtained by adding this vector to the eye coordinates is considered the 3D gaze target.
    \item \emph{Hu et al.}~\cite{hu2023gfie} is the most relevant work, which uses head crops, scene images (without blurring), and depth maps to build a point cloud and simultaneously estimate the gaze direction and location in both 2D and 3D.
\end{itemize}

The baseline results, obtained with the \emph{Random} and \emph{Center} heuristics, do not offer particular insights about the problem, achieving comprehensively high error values in both scenarios. 
The methods built explicitly for gaze estimation (\mbox{\emph{Rt-Gene}~\cite{fischer2018rt}} and \emph{Gaze360}~\cite{kellnhofer2019gaze360}) perform well in estimating a suitable gaze vector, while obtaining poor results on the gaze target selection. 

Our methodology, trained with a multi-stage strategy or an end-to-end strategy and based on a gaze estimation method that exploits the attention mechanism on the feature extracted from both upper-body poses and depth maps and operates on scene images with the faces blurred out, outperforms all state-of-the-art in three metrics out of four. More importantly, both the 3D metrics have been improved with respect to~\cite{hu2023gfie}, with a gain of 27 millimeters on the 3D distance and a 1.8\degree~improvement on the 3D angle error achieved by the multi-stage strategy, with the gaze estimation module trained alone before the full pipeline training. 
Improving gaze estimation is, therefore, of fundamental importance in solving the task of gaze target detection in 3D, even more than in 2D scenarios. 
Since most of the models' inputs are sensory-acquired 3D data (depth maps, camera parameters, and eye positions), a better estimation of the gaze vector would allow focusing on a relatively small number of potential 3D targets.

\begin{table}[!tb]
  \caption{Results on the GFIE dataset~\cite{hu2023gfie} for the 2D and the 3D scenarios. The results of the prior work are taken from~\cite{hu2023gfie}. $Dist.$ stands for Distance. 
  The \textbf{bold} numbers represent the best results. The second best is \underline{underlined}. 
  H, D, UP, and S stand for head crop, depth map, upper-body pose, and scene, respectively. 
  }
  \label{tab:results}
  \centering
  \resizebox{0.9\textwidth}{!}{%
  \begin{tabular}{lcccccc}
    \toprule
    & \textbf{Original} & \textbf{Privacy} \hspace{1mm} &\multicolumn{2}{c}{\underline{\textbf{3D Metrics}}} & \multicolumn{2}{c}{\underline{\textbf{2D Metrics}}} \\
    \textbf{Method} & \textbf{Modalities} & \textbf{Preserving} \hspace{1mm} & \textbf{Dist.(↓)} & \textbf{Angle Error(↓)} & \textbf{AUC(↑)} & \textbf{Dist.(↓)} \\
    \midrule
    Random & S & \xmark & 2.930 & 84.4\degree & 0.585 & 0.425 \\ 
    Center & S & \xmark & 2.510 & 87.2\degree & 0.614 & 0.287 \\
    GazeFollow~\cite{recasens2015they} & H, S & \xmark & 0.856 & 41.5\degree & 0.941 & 0.131 \\
    Lian et al.~\cite{lian2018believe} & H, S & \xmark & 0.542 & 26.7\degree & 0.962 & 0.091 \\
    Chong et al.~\cite{chong2020detecting} & H, S & \xmark & 0.455 & 20.8\degree & 0.972 & \underline{0.069} \\
    Rt-Gene~\cite{fischer2018rt} & H, S & \xmark & 0.552 & 21.0\degree & 0.823 & 0.123 \\
    Gaze360~\cite{kellnhofer2019gaze360} & H, S & \xmark & 0.540 & 19.8\degree & 0.821 & 0.130 \\
    Hu et al.~\cite{hu2023gfie} & H, S, D & \xmark & 0.311 & 17.7\degree & 0.965 & \textbf{0.065} \\
    \textbf{Ours (Multi-stage)} & UP, D, S (blurred) & \cmark & \textbf{0.284} & \textbf{15.9\degree} & \textbf{0.983} & 0.083 \\ 
    \textbf{Ours (End-to-end)} & UP, D, S (blurred) & \cmark & \underline{0.292} & \underline{16.2\degree} & \underline{0.973} & 0.083
    \\
  \bottomrule
  \end{tabular}}
\end{table}

\subsection{Qualitative Results}
\label{subsec:qual}

\begin{figure}[!tb]
  \centering
  \includegraphics[width=.85\linewidth]{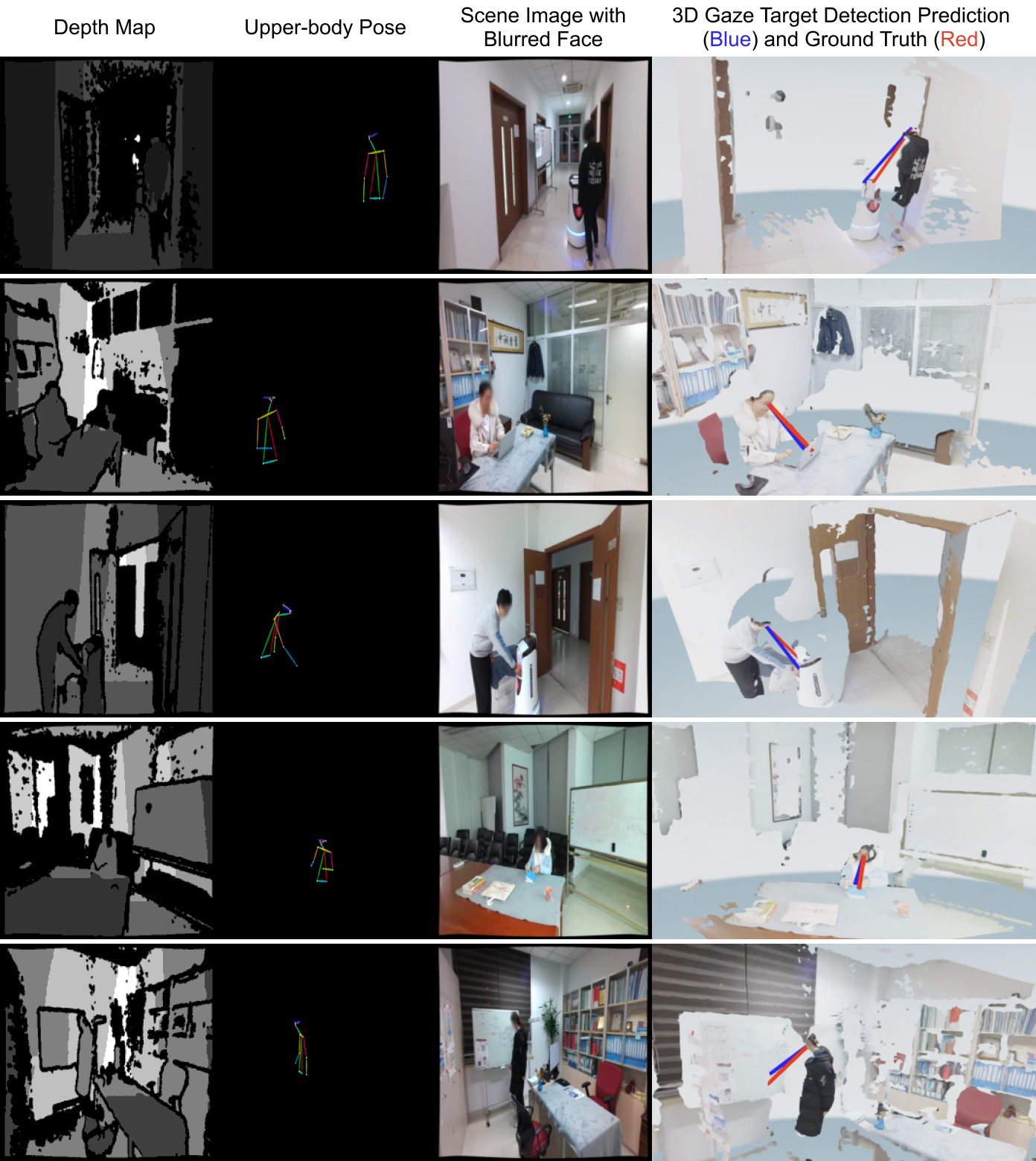}
  \caption{ 
  Qualitative results of our method on the GFIE dataset~\cite{hu2023gfie}. Each row represents a single sample. \textbf{First Column:} We use resized depth maps as input for our gaze estimation module. \textbf{Second Column:} The estimated upper-body skeletons are visualized on a black background before being normalized. \textbf{Third Column:} Scene images where the faces are blurred out before being used by the encoder-decoder module to predict the 2D gaze target heatmap. \textbf{Fourth Column:} The final point cloud of the scene, with the ground truth gaze vector in {\color{red}red} and the estimated gaze vector in {\color{blue}blue}.
  }
  \label{fig:gfie-qualitatives}
\end{figure}

In Fig.~\ref{fig:gfie-qualitatives}, we present some qualitative results obtained on the GFIE dataset~\cite{hu2023gfie}. As it can be seen from the analyzed samples, our method can estimate 3D gaze vectors and gaze targets (in {\color{blue}blue}) very close to the ground truth annotated ones (in {\color{red}red}). More interestingly, our gaze estimation module seems to operate well regardless of the orientation of the person, with upper-body skeletons front facing the camera (fourth row), on the side (fifth row), or captured from the back (first row). Additionally, in the third row, a subject interacts with a robot, both looking and touching it, demonstrating the relationship between body position, the performed action, and, consequently, where the gaze is directed.

\begin{table}[!tb]
  \caption{Ablation study in terms of modality and the type of input scene image (\ie, face blurred and not blurred) on the GFIE dataset~\cite{hu2023gfie}. $Dist.$ stands for Distance. Note that depth maps and scene images are fundamental and must be included for the 3D gaze target detection to function properly; thus, their removal cannot be ablated. 
  }
  \label{tab:ablation}
  \centering
  \resizebox{0.9\textwidth}{!}{%
  \begin{tabular}{lcccccccc}
    \toprule
   & \textbf{Head} \hspace{1mm} &\textbf{Blurred} \hspace{1mm} & \textbf{Full} \hspace{1mm} & \textbf{Upper} \hspace{1mm} &\multicolumn{2}{c}{\underline{\textbf{3D Metrics}}} & \multicolumn{2}{c}{\underline{\textbf{2D Metrics}}} \\
   & \textbf{Crop} \hspace{1mm} & \textbf{Face} \hspace{1mm} & \textbf{Body} \hspace{1mm} & \textbf{Body} \hspace{1mm} & \textbf{Dist.(↓)} & \textbf{Angle Error(↓)} & \textbf{AUC(↑)} & \textbf{Dist.(↓)} \\
    \midrule
        & \cmark & \xmark  & \xmark & \xmark & 0.311 & 17.7\degree & 0.965 & 0.065 \\
        & \xmark & \xmark  & \xmark & \cmark & 0.314 & 16.1\degree & 0.981 & 0.086 \\
        & \xmark & \cmark  & \cmark & \xmark & 0.310 & 16.7\degree & 0.979 & \textbf{0.059} \\
     Multi-stage & \xmark & \cmark  & \xmark & \cmark & \textbf{0.284} & \textbf{15.9\degree} & \textbf{0.983} & 0.083 \\
     End-to-end & \xmark & \cmark  & \xmark & \cmark & 0.292 & 16.2\degree & 0.973 & 0.083 \\
  \bottomrule
  \end{tabular}}
\end{table}

\subsection{Ablation Study}
\label{subsec:ablation}
We report ablation results (Table~\ref{tab:ablation}) performed by removing or changing the used modalities for the proposed method. 
We also present the results of the proposed method when trained using either the end-to-end approach or the multi-stage training process. 
As seen, using the upper-body skeleton instead of the complete 17-joint skeleton is slightly beneficial for the 3D Distance compared to the baseline (third row). 
Interestingly, using the entire skeleton leads to the best 2D distance score. 
This can happen due to the optimization process of the two losses, $\mathcal{L}_{gaze}$ and $\mathcal{L}_{heat}$. 
In this case, it seems the latter was favored over the former.
The second row shows that using the model without blurring the faces benefits the 2D AUC metric.
The pipeline trained with a multi-stage policy, firstly training the gaze estimation module alone and loading its weights before launching the training on the full pipeline, achieved the best results on both the 3D metrics, with a minimum of 15.9\degree~for the angle error. 
However, if a faster training procedure is required, the end-to-end strategy still provides outstanding results and is preferable in such a situation.
\section{Conclusion}
\label{sec:conclusion}
The gaze target detection task, often referred to as gaze-following, is a fundamental proxy task for multiple vision problems regarding the behavior and interaction of people between themselves and with the environment. 
Understanding where and what a person is observing can give us hints about their current interests and future actions, whether they are interactions with humans, with objects in the environment, and, more recently, with collaborative robots that are increasingly present in public and workplaces.
In this paper, we propose a novel way to tackle 3D gaze estimation and 3D gaze target detection by focusing on the upper-body pose and the scene depth map, outperforming the previous state-of-the-art.
The combination of the skeleton-based gaze estimation module and the blurring applied on the face in the scene image allowed us to present an effective method that is also able to preserve the sensitive data of the person, such as the face appearance, generally used to solve these types of tasks.

\noindent\textbf{Limitations and future work.}
In instances where pose estimation fails, our upper-body-based gaze estimation and target detection pipeline might not perform as effectively. However, this issue is also present in state-of-the-art methods that primarily rely on head crops. For the dataset used in this paper, we manually reviewed the body poses for a subset of the testing data and did not observe any failures with the pose estimation method employed. At present, there is a lack of multi-person 3D gaze target detection datasets. Consequently, our work is constrained by the complexity of scenes and the availability of a dataset that only features single individuals per scene. As more crowded datasets and corresponding gaze annotations become available, it will be important to test our approach on these datasets as well. However, we currently anticipate that our pipeline can be applied directly without requiring modifications. Eventually, in case of occlusions across multiple persons, we can adopt a denoising pose estimator to our pipeline. Furthermore, an interesting future direction is the inclusion of activity recognition as part of the pipeline, exploiting the gaze direction and target as a proxy for predicting the future intentions of one or more people.

\section*{Acknowledgements}
This study was carried out within the PNRR research activities of the consortium iNEST (Interconnected North-Est Innovation Ecosystem) funded by the European Union Next-GenerationEU (Piano Nazionale di Ripresa e Resilienza (PNRR) – Missione 4 Componente 2, Investimento 1.5 – D.D. 1058  23/06/2022, ECS\_00000043).
This manuscript reflects only the Authors’ views and opinions. Neither the European Union nor the European Commission can be considered responsible for them.

%
%
\bibliographystyle{splncs04}
\bibliography{bibi}
\end{document}